\documentclass[final]{cvpr}

\usepackage{times}
\usepackage{epsfig}
\usepackage{graphicx}
\usepackage{amsmath}
\usepackage{amssymb}

\usepackage[numbers,sort,compress]{natbib}
\usepackage{color,soul}
\usepackage{booktabs}
\usepackage{multirow}
\usepackage{tabularx}
\usepackage{mathabx,bm}
\usepackage[table,xcdraw]{xcolor}

\usepackage[pagebackref=true,breaklinks=true,colorlinks,bookmarks=false,citecolor=red]{hyperref}

\def\cvprPaperID{6933} % *** Enter the CVPR Paper ID here
\def\confYear{CVPR 2021}
\def\ie{\textit{i.e.}, }
\def\eg{\textit{e.g.}, }
\def\etc{\textit{etc.} }
\begin{document}

%%%%%%%%% TITLE
\title{Large-Scale Generative Data-Free Distillation}

\author{Liangchen Luo\thanks{Work done as part of the Google AI Residency.}, Mark Sandler, Zi Lin\footnotemark[1], Andrey Zhmoginov, Andrew Howard\\
Google Research\\
{\tt\small \{luolc,sandler,lzi,azhmogin,howarda\}@google.com}
% For a paper whose authors are all at the same institution,
% omit the following lines up until the closing ``}''.
% Additional authors and addresses can be added with ``\and'',
% just like the second author.
% To save space, use either the email address or home page, not both
}

\maketitle

%%%%%%%%% ABSTRACT
\begin{abstract}
Knowledge distillation is one of the most popular and effective techniques for knowledge transfer, model compression and semi-supervised learning. 
Most existing distillation approaches require the access to original or augmented training samples. 
But this can be problematic in practice due to privacy, proprietary and availability concerns. 
Recent work has put forward some methods to tackle this problem, but they are either highly time-consuming or unable to scale to large datasets. 
To this end, we propose a new method to train a generative image model by leveraging the intrinsic normalization layers' statistics of the trained teacher network. 
This enables us to build an ensemble of generators without training data that can efficiently produce substitute inputs for subsequent distillation.
The proposed method pushes forward the data-free distillation performance on CIFAR-10 and CIFAR-100 to $95.02\%$ and $77.02\%$ respectively. 
Furthermore, we are able to scale it to ImageNet dataset, which to the best of our knowledge, has never been done using generative models in a data-free setting.
\end{abstract}

%%%%%%%%% BODY TEXT
\section{Introduction}
\newcommand{\omt}[1]{}
\omt{Recent advances in representation learning have demonstrated the powerful role played by deep a neural networks. 
As a result,} 
Recent advances in deep learning have dramatically accelerated machine learning progress in a wide variety of artificial cognition tasks, including vision~\cite{Simonyan2015VeryDeepConv,Krizhevsky2012AlexNet,Ren2015FasterRCNN,Long2015FullyConvSeg}, speech~\cite{graves2013speech} 
and natural language processing~\cite{NLPDeep2020}.  
%detection~\cite{Ren2015FasterRCNN}, and semantic segmentation~\cite{Long2015FullyConvSeg}. % \needcite{more?}
One intrinsic property of deep neural networks is that they constitute a black box, with no interpretable way of describing how they perform tasks as they are trained for. While there has been significant progress in understanding how feature maps represent human interpretable features \cite{Mordvintsev2015DeepDream,kim2018interpretability}, we still lack a complete picture of how neural networks operate. 
This limits our ability to extend and modify existing neural networks.  Recently knowledge distillation \cite{romero2014fitnets, Hinton2014KD} has emerged as a robust way of transferring knowledge across different neural networks~\cite{crowley2018moonshine,hahn2019self,Wang2020-lh}. 
In the teacher-student framework, the student is taught to mimic the prediction layer of the teacher.
The knowledge distillation approach has been an instrumental technique for model compression~\cite{crowley2018moonshine,hahn2019self}, as well as for pushing state of the art by harnessing large amounts of unlabeled data to improve the performance of student models such as Noisy Student \cite{xie2020selftraining} and many others, see~\cite{Wang2020-lh} for a survey. 

Despite its successes, knowledge distillation in its classical form has a critical limitation. It assumes that the real training data is still available in the distillation phase. However, in practice, the original training data is often unavailable due to privacy concerns. 
A typical example \omt{of such availability concerns} is medical imaging records~\cite{Frid2018GANMedImageAug,Han2019GANMedImageAug}, whose availability is time-limited due to security and confidentiality.
% One might be able to train a model on such data but later find it already expired when he wants to do distillation.
% \notsure{Another common pitfall comes from the size of data.}
Similarly many large models are trained on millions~\cite{Deng2009ImageNet} or even billions~\cite{Mahajan2018InsHashTag} of samples.
While the pre-trained models might be made available for the community at large, making training data available poses a lot of technical and policy challenges. Even storing the training data is often infeasible for all but the largest users. Furthermore, in some training frameworks such as federated learning~\cite{Konecny2016FederatedLearning}, the model is tuned by collecting gradient updates calculated on distributed devices, and the training data disappears as soon as its been used, thus requiring even more data each time the model needs to be updated. 

\begin{figure}[tb]
    \centering
    \includegraphics[width=\linewidth]{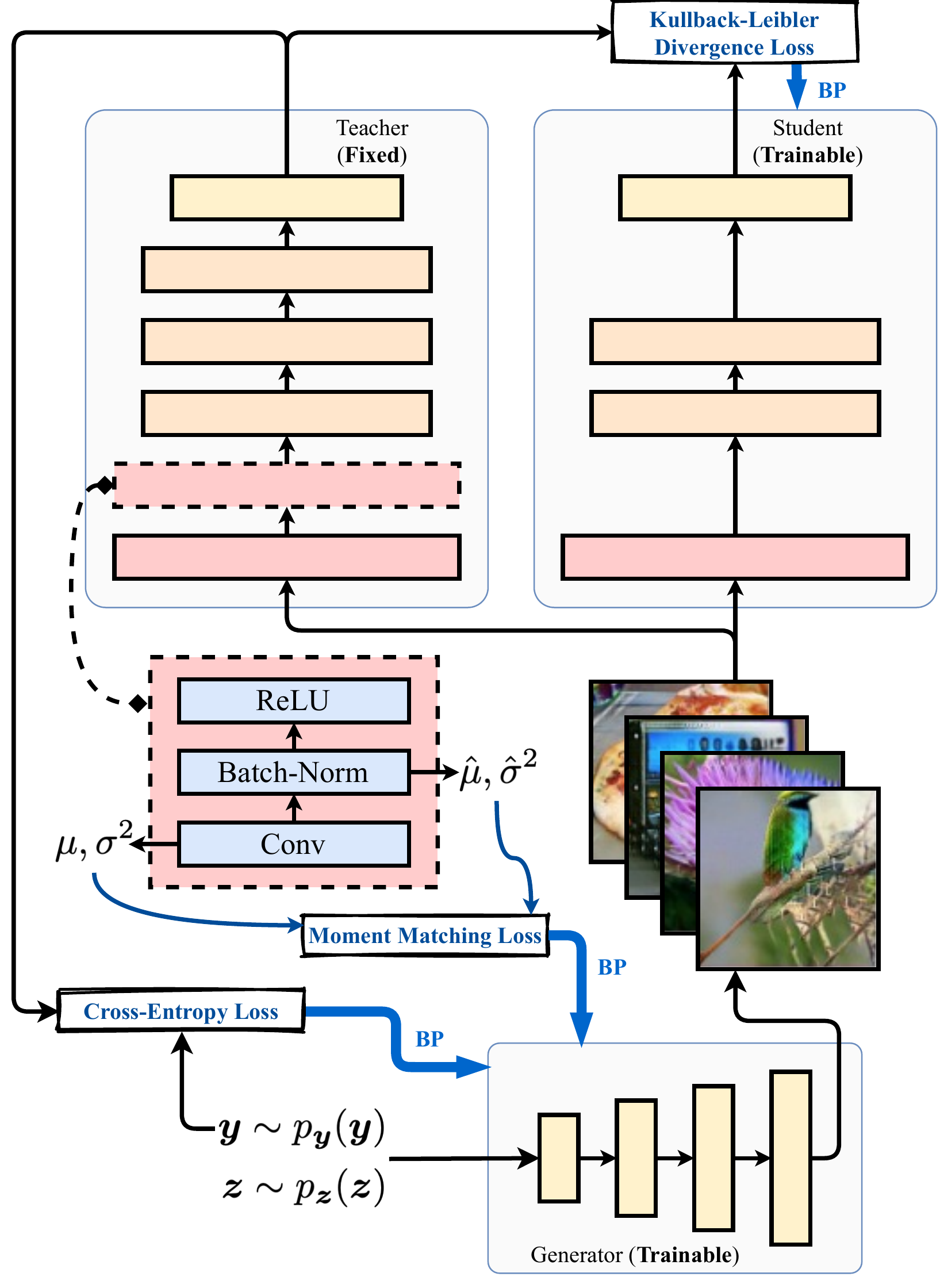}
    \caption{
        The proposed generative data-free distillation method.
        The generator is trained without the real images by (1) maximizing the probability of a target label being predicted by the pre-trained teacher; and (2) matching the statistics ($\mu$ and $\sigma^2$) of the batch-norm layers (see Eq.~\eqref{eq:stat-matching}).
        Subsequently, the synthetic images produced by the generator enables us to apply knowledge distillation.
        More examples of generated images are shown in Figure~\ref{fig:imagenet-visual}.
    }
    \label{fig:model}
\end{figure}

To address this issue, a few approaches to the data-free knowledge distillation, \ie for distilling models under a data-free regime, have been previously proposed. 
However, they are either highly time-consuming to produce synthetic images or unable to scale to large datasets.
We disciss prior research in this field in Section~\ref{sec:related-work}.

In this work, we adapt the idea of generative image modeling to attain efficient data generation, and investigate ways to scale it to large datasets.
We propose our generative data-free distillation method, as illustrated in Figure~\ref{fig:model}, by training a generator without using the original training data and use it to produce substitute data for knowledge distillation. 
Our generator minimize two optimization objectives: 
(1)~\textit{moment matching loss}, in which the generator minimizes the difference between activation statistics and known moments estimated on training data; and 
(2)~\textit{inceptionism loss}, in which the generator maximizes the activation of the logit of the teacher network corresponding to the target loss.  
The variants of moment-matching loss has been explored before in non-generative data-free image synthesis methods such as in \cite{Yin2020DeepInversion,Haroush2020StatMatchQuant}. We also note that this information is often available as part of the training batch-normalization~\cite{Ioffe2015BN} layers which are present in nearly all modern architectures such as ResNets~\cite{He2016ResNet}, DenseNets~\cite{Huang2017DenseNet}, MobileNets~\cite{Howard2017MobileNetV1} and their variants.
%In other words, they store the statistics of the training images at different levels of representation.
Under an isotropic Gaussian assumption of the internal activations, we can explicitly minimize their Kullback--Leibler divergence or the $\ell_2$-norm of their difference. %, which is called \textit{moment matching loss} in this paper.

We then follow the idea of deep dream style \cite{Mordvintsev2015DeepDream} image synthesis method to employ \textit{inceptionism loss}.
The general idea is to find an input image that can maximize the probability of a certain category being predicted by the pre-trained teacher, which can be naturally formulated as a cross-entropy minimization problem.
Combining this with the aforementioned moment matching loss together, given only a pre-trained teacher model, we are now able to train a generator without using real images, which can effectively produce synthetic images for the distillation.

% We then propose our generative data-free distillation method, illustrated in Figure~\ref{fig:model}, by introducing this sort of moment matching loss together with the inceptionism~\cite{Mordvintsev2015DeepDream} loss to generative image modeling.
% This enables us to train an image generator without access to the original training data.
% The generator would be able to produce substitute data efficiently for the subsequent knowledge distillation process.
% In addition, using the ensemble of multiple generators demonstrates its ability of further improving the distillation result as well as scaling to a larger and more complex dataset.

To demonstrate the effectiveness of the proposed method, we design an empirical study on three image classification datasets with increasing size and complexity.
We first conduct an experiment of data-free distillation on CIFAR-10 and CIFAR-100.
The generator trained without using real images is able to produce higher-quality and more realistic images than previous methods.
These images can also effectively support the following knowledge distillation. The learned student outperforms the previous methods with a clear margin, achieving a new state-of-the-art result which is even better than its supervised-trained counterparts.
We then explore using the ensemble of multiple generators on CIFAR-100 and ImageNet and demonstrate its ability of further improving the distillation result. 

% We first illustrate the specific contributions of different loss objectives we introduce to the generative image model, by training with each objective and evaluate its subsequent distillation performance on CIFAR-10 independently.
% Afterward, we show that the generator trained on these objectives together can produce higher-quality and more realistic images than previous methods.

% We then move on to CIFAR-100 and also show better performance using our method comparing with existing works.
% Besides, we conduct experiments of distillation with the ensemble of $100$ generators, which can further improve the result by $\sim 0.5\%$ in terms of the accuracy score.
% Finally, we turn to ImageNet to show how our method scale to a larger dataset and achieve comparable results with costly non-generative methods.
%as well as scaling to a larger and more complex dataset.

Our main contributions can be summarized as follows:
\begin{itemize}
\setlength\itemsep{0em}
  \item We propose a new method for training an image generator from a pre-trained teacher model, which efficiently produces synthetic inputs for knowledge distillation.
  \item We push forward the state of the art of data-free distillation on CIFAR-10 and CIFAR-100 datasets to $95.02\%$ and $77.02\%$ respectively, which is even better than the supervised-trained counterparts.
  \item We scale the generative data-free distillation method to ImageNet by using multiple generators. This is the first success of data-free distillation on ImageNet using generative models to the best of our knowledge.
\end{itemize}
\section{Related Work}
\label{sec:related-work}

Our approach can be viewed as a combination of two components, \textit{generative modeling} and \textit{knowledge distillation}, each of which attracted considerable attention of the deep learning community over the past years.
Here we provide a brief overview of these fields in the context of our work.

% =================================================================================
% Generative Image Modeling
% =================================================================================
\textbf{Generative Image Modeling.}
Generative Adversarial Networks (GANs) \cite{Goodfellow2014GAN}, perhaps the most celebrated approach to image generation, together with numerous variants of this general method \cite{Brock2019BigGAN,Arjovsky2017-as,Karras2017-vv}  have shown a tremendous potential for generating high-fidelity synthetic images based on a limited corpus of training samples.
One appealing application of synthetic image generation is data augmentation.
Recent studies have employed GAN-based augmentation to improve the model performance in data-restricted scenarios such as medical imaging \cite{Frid2018GANMedImageAug,Han2019GANMedImageAug}.
However, as stated in \citet{Goodfellow2019AAAITalk}, this approach has not shown much success in practice on large-scale data.
It was also observed that while the image are of extremely high quality, using them exclusively for training leads to a significant performance degradation~\cite{Yin2020DeepInversion}.

Another promising approach to image synthesis is based on recent work on reversible networks \cite{Behrmann2019iResNet,Gomez2017RevNet,Jacobsen2018iRevNet}.
These studies explore reversible models, in which the transformation from one layer to the next are invertible, allowing to reconstruct layer activations using the outputs of the following layer.
The initial motivation was to save memory by computing the activations on-the-fly during backpropagation, while later researchers have also discovered its potential for image generation~\cite{Asim2020iVNetDenoise}. 
We stress here, that while GAN and GAN-like methods generate very realistic and high-quality images, all aforementioned methods require access to the original data to build  their generators.

% =================================================================================
% Knowledge Distillation
% =================================================================================
\textbf{Knowledge Distillation.}
% Knowledge distillation is a knowledge transfer technique that can be used for network compression in which a compact model, the \textit{student}, is trained under the guidance of a larger model, the \textit{teacher}, or an ensemble of such models.
% In a typical image classification task, the logits, \eg the outputs of last layer of a deep neural network, are used as %the carriers of the knowledge from the teacher model, beyond what can be extracted from the training set alone.
% In principle, a model performing well on the training set is likely to provide confident predictions, \ie output high %probability for the correct class and near-zero probabilities for all other classes.
% But it is a general intuition that these ``non-zero'' probabilities carry additional information about the teacher %model itself and, to some extent, describe generalization capabilities of the model and how well it will perform on the %test set.
Knowledge distillation \cite{Hinton2014KD} is a general technique that can be used for model compression.
It transfers knowledge from a pre-trained network, the \textit{teacher}, to another \textit{student} network by teaching the student to mimic the teacher's behaviour.
In a typical image classification task, this is usually done by aligning the probabilities predicted by the teacher and student network.
In recent literature, there have been numerous variations of knowledge distillation in terms of application domains and distillation strategies. For an overview of general distillation techniques we refer the reader to knowledge distillation surveys such as \cite{Wang2020-lh}.

% One direction is to transfer additional knowledge in the form of intermediate activations \cite{romero2014fitnets,aguilar2020knowledge,zhang2017knowledge}, attention maps \cite{zagoruyko2016paying}, weight projections \cite{lee2018self} or layer interactions \cite{yim2017gift}.
% Other methods try to directly address the capacity gap between a teacher and student by distilling from a series of intermediate teachers \cite{mirzadeh2020improved,jin2019knowledge}. \todo{comments?}

% =================================================================================
% Data-Free Model Compression
% =================================================================================
\textbf{Data-Free Knowledge Distillation.}
% The problem of knowledge distillation under a special constraint that the original training data is not available during distillation has attracted a growing interest in privacy-sensitive and data-restricted scenarios.
The problem of knowledge distillation becomes much more challenging when the original training data is not available at the time to train the student model. 
This is often encountered in privacy-sensitive and data-restricted scenarios. 
% The vast majority of methods in this setting essentially resort to synthetic image generation.
Most approaches to this scenario center around synthetic image generation.
\citet{lopes2017data} was a first attempt to pre-compute and store activation statistics for each layer of the teacher network with the goal of constructing synthetic inputs that produces similar activations.
Follow-up works~\cite{nayak2019zero,Yin2020DeepInversion} have developed the approach by using less meta-data or proposing different optimization objectives.
These methods typically obtain the synthetic inputs by directly optimizing some trainable random noise with regards to a pre-determined objective, where each input image requires multiple iterations to converge.
Therefore, it can be costly and time-consuming to produce sufficient data for compression.
There are also a few methods that synthesize input data via generative image modeling \cite{Chen2019DAFL,Fang2019DFAD,micaelli2019zero}, which create substitute data much more efficiently than optimizing input noise.
However, scaling them to the tasks on large datasets, \eg ImageNet classification task, remains challenging.

% Our work follows the track of generative image modeling to attain efficient data generation, and \notsure{is the first that can successfully be applied to ImageNet to the best of our knowledge.}

\section{Generative Distillation in Data-Free Setting}

In this section, we first briefly recall the classical knowledge distillation method and then introduce our approach for building a generative model from a pre-trained teacher.

\subsection{Notation}

Generally, we denote random variables with bold serif font, \eg $\bm{x}$, $\bm{z}$.
By contrast, sampled values of such variables and deterministic tensors are denoted with regular serif font, \eg $x$, which is typically used when we are discussing loss objectives with respect to a single input.
Loss functions denoted by $\mathcal{L}_{\blacktriangle}$, where $\blacktriangle$ is an abbreviation of a particular loss component, typically involve averaging over the probability distributions entering them.
Without ambiguity, we may slightly abuse the notation of $\mathcal{L}_{\blacktriangle}(x)$ to denote the deterministic loss value with respect to a single input $x$.

% =================================================================================
% Knowledge Distillation
% =================================================================================
\subsection{Knowledge Distillation}
Knowledge distillation aims to transfer knowledge from typically a larger teacher network $T(\bm{x}; \theta_t)$ into a smaller student $S(\bm{x}; \theta_s)$, where $T$ and $S$ are usually differentiable functions represented by neural networks with parameters $\theta_t$ and $\theta_s$ respectively and $\bm{x}$ is the model input.
In the setting of a classification task, $T$ and $S$ typically output a probability distribution over~$K$ different possible categories.
% The student network is trained to mimic the behavior of teacher network by matching the temperature-scaled soft target distribution produced by the teacher on training images.
The student is trained to mimic the behavior of the teacher network by matching the probability distribution produced by the teacher on the training data.
% Let $\bm{p} = T(\bm{x})$ and $\bm{q} = S(\bm{x})$ represent the distribution outputs of the teacher and the student networks respectively.
% Let $T(\bm{x})$ and $S(\bm{x})$ represent the distribution outputs of the teacher and the student networks respectively.
% Then formally,
Formally, knowledge distillation can be modeled as a minimization of the following objective:
\begin{equation}
\label{eq:KD}
    \mathcal{L}_\text{KD}= \mathbb{E}_{\bm{x} \sim
    % p_\text{data}(\bm{x})}[D_\text{KL}(\bm{p} \,\|\, \bm{q})],
    p_\text{data}(\bm{x})} \big[ D_\text{KL}\big(T(\bm{x}) \,\|\, S(\bm{x}) \big) \big],
\end{equation}
where $D_\text{KL}(\cdot \| \cdot)$ refers to the Kullback--Leibler divergence that evaluates the discrepancy between the distributions produced by the teacher and student networks.
Here $p_\text{data}$ denotes the training data distribution.

% =================================================================================
% Generator Training
% =================================================================================
\subsection{Generative Image Modeling}

Computing the loss objective in Equation~\eqref{eq:KD} requires the knowledge of the data distribution $p_\text{data}$, which is not available in a data-free setting.
Instead, we approximate $p_\text{data}(\bm{x})$ with a distribution of a generator trained to mimic the original data.
We learn the generator distribution $p_g(\tilde{\bm{x}} \,|\, \bm{y})$ conditioned on the class label $\bm{y}$ given the trained teacher $T$ by introducing a latent variable $\bm{z}$ with a prior $p_{\bm{z}}(\bm{z})$ and representing $p_g$ as a marginal $\mathbb{E}_{\bm{z}\sim p_{\bm{z}}} \delta(\tilde{\bm{x}}-G(\bm{z} \,|\, \bm{y}))$, essentially learning a deterministic deep generator $G(\bm{z} \,|\, \bm{y}; \theta_g)$.
%The generator $G(\bm{z},\bm{y};\theta_g)$ is chosen to be a GAN model conditioned on the class label $\bm{y}$ with the latent variable $\bm{z}$ defined by a prior distribution $p_{\bm{z}}(\bm{z})$.
The generator is then trained without the access to $p_\text{data}$, but using only the trained teacher model $T$.
% and some additional pre-computed data.
Now the key is to find appropriate objectives for training the generator.
These objectives are introduced in the remainder of this section.

% =================================================================================
% Inceptionism (DeepDream)
% =================================================================================
\textbf{Inceptionism loss.}
Inceptionism-style \cite{Mordvintsev2015DeepDream} image synthesis, also known as DeepDream, is a way to visualize input images that provoke a particular response of a trained neural network.
For instance, say we want to know what kind of image would result in ``dog'' class being predicted by the model.
The inceptionism method would start with a trainable image $x$ initialized with random noise, and then gradually tweak it towards the most ``dog-like'' image by maximizing the probability of the dog category being produced by the model.
Formally, given the expected label $\hat{y}$ and the trained teacher $T$, we find $x$ that minimizes the cross-entropy of the categorical distribution $\hat{p} = \operatorname{OneHot}(\hat{y})$ relative to $p = T(x)$:
\begin{equation}
\label{eq:cross-entropy}
    \mathcal{L}_\text{CE}(x, \hat{y}) = H(\hat{p}, p) = - \sum_i \hat{p}_i \log p_i .
\end{equation}
In practice we usually do not optimize this objective alone, but also impose a prior constraint that the synthetic images mimic the statistics to the natural images, such as a particular correlation of neighboring pixels.
It is done by adding a regularization term to Equation~\eqref{eq:cross-entropy}:
\begin{equation}
    \mathcal{L}_\text{Inc}(x, \hat{y}) = \mathcal{L}_\text{CE}(x, \hat{y}) + \mathcal{L}_\text{Reg}(x),
\end{equation}
where in this paper we follow \cite{Yin2020DeepInversion,Haroush2020StatMatchQuant} to use total variation loss and $\ell_2$-norm loss as the regularizers:
\begin{equation}
    \mathcal{L}_\text{Reg}(x) = \lambda_\text{t} \mathcal{L}_\text{t}(x) + \lambda_{\ell_2} \mathcal{L}_{\ell_2}(x),
\end{equation}
where $\mathcal{L}_\text{t}$ and $\mathcal{L}_{\ell_2}$ penalize the total variation and $\ell_2$-norm of the image with scaling weights $\lambda_\text{t}$ and $\lambda_{\ell_2}$, respectively.

% =================================================================================
% Moment Matching Loss (Batch-Norm Statistic Matching Loss)
% =================================================================================
\textbf{Moment matching loss.}
% \todo{I would prefer to avoid inventing new terms and call this a ``moment-matching'' loss.}
The inceptionism loss by itself only constraints the input (images) and output (probabilities) of the trained network, while leaving the activations of internal layers unconstrained.
Previous studies have observed that different layers of a deep convolutional network are likely to perform different tasks \cite{Lee2009Conv,Luo2019AdaBound,Gu2018ConvAdavances}, \ie lower layers tend to detect low-level features such as edges and curves, while higher layers learn to encode more abstract features.
In addition, \citet{Haroush2020StatMatchQuant} showed that images learned with conventional inceptionism method may result in anomalous internal activations deviating from those observed for real data.
These facts suggest there should be a regularization term to constrain the statistics of the teacher's intermediate layers as well.

Batch normalization \cite{Ioffe2015BN} layers, a common component of most neural networks, are helpful with providing such statistics \cite{Yin2020DeepInversion,Haroush2020StatMatchQuant}. The normalization operation is designed to normalize layer activations by re-centering and re-scaling them with the moving averaged mean and variance calculated during training.
In other words, it implicitly stores the estimated layer statistics of the teacher over the original data $p_\text{data}(\bm{x})$.
Therefore, we can force the layer statistics (mean and variance specifically) produced by our synthetic images to align with those emerging from the real data \cite{Salimans2016,WardeFarley2017,Nogueira2019}. % \needcite{moment matching.}

Given the running estimates for the mean $\hat{\mu}$ and the variance $\hat{\sigma}^2$ of a teacher batch-norm layer, we measure the mean~$\mu(x)$ and variance~$\sigma^2(x)$ activated by the synthetic image~$x$ and minimize the discrepancy between the real-time statistics and the estimated ones.
Under an isotropic Gaussian assumption, this can be done by minimizing their Kullback--Leibler divergence:
\begin{multline}
\label{eq:stat-matching-KL}
    D_\text{KL} \big( \mathcal{N}(\hat{\mu}, \hat{\sigma}^2) \,\|\, \mathcal{N}(\mu, \sigma^2) \big)
     \\ = \log \frac{\sigma}{\hat{\sigma}} - \frac{1}{2} \Big[ 1 - \frac{\hat{\sigma}^2 + (\mu - \hat{\mu})^2}{\sigma^2} \Big],
\end{multline}
or the $\ell_2$-norm of their differences:
\begin{equation}
\label{eq:stat-matching-l2}
    \lVert \mu - \hat{\mu} \rVert_2 + \lVert \sigma^2 - \hat{\sigma}^2 \rVert_2,
\end{equation}
where $\mathcal{N}(\cdot,\cdot)$ stands for Gaussian distribution and $\lVert \cdot \rVert_2$ denotes the $\ell_2$-norm.
% In this paper, we choose the latter one and formulate the \textit{statistic matching loss} by summing up these penalties together across all the batch-norm layers:
In this paper, we choose the latter one and formulate the \textit{moment matching loss} by summing up these penalties together across all the batch-norm layers:
\begin{equation}
\label{eq:stat-matching}
    \mathcal{L}_\text{M}(x) = \lambda_\text{S} \sum_l \big[ \lVert \mu_l(x) - \hat{\mu}_l \rVert_2 + \lVert \sigma^2_l(x) - \hat{\sigma}^2_l \rVert_2 \big],
\end{equation}
where $\lambda_\text{S}$ is the scaling weight and $\hat{\mu}_l$, $\hat{\sigma}^2_l$ are recorded mean and variance statistics 
for per-layer activations on the real data.

% =================================================================================
% Generator Training loss
% =================================================================================
\textbf{Generator training objective.}
Combining the inceptionism loss and moment matching loss together we get
\begin{equation}
\label{eq:img-loss}
    \mathcal{L}_\text{Image}(x, y) = \mathcal{L}_\text{Inc}(x, y) + \mathcal{L}_\text{M}(x).
\end{equation}
Recall that our final goal is to utilize these objectives to train a generative model.
By substituting $x$ with $G(\bm{z} \,|\, \bm{y})$ in Equation~\eqref{eq:img-loss}, we define the final generator training objective as:
\begin{equation}
\label{eq:g-loss}
    \mathcal{L}_\text{G} = \mathbb{E}_{\bm{z} \sim p_{\bm{z}}(\bm{z}), \bm{y} \sim p_{\bm{y}}(\bm{y})} \big[ \mathcal{L}_\text{Inc} \big( G(\bm{z} \, | \, \bm{y}), \bm{y} \big) + \mathcal{L}_\text{M} \big( G(\bm{z} \, | \, \bm{y}) \big) \big].
\end{equation}

\begin{figure}[htb]
    \centering
    \includegraphics[width=0.8\linewidth]{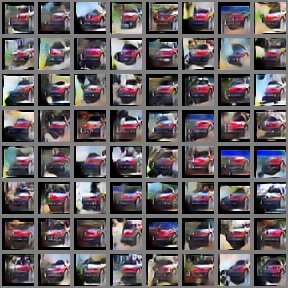}
    \caption{
        An example of mode collapse happened to the generator trained in data-free setting.
        The objects in generated images in class ``automobile'' are all in red color.
    }
    \label{fig:cifar10-mode-collapse}
\end{figure}

\textbf{Using multiple generators.} 
{\em Mode collapse} is a common problem plaguing various generative models like GANs \cite{Mao2019MSGANModeCollapose,Goodfellow2014GAN,Salimans2016ImprovedGAN}, where instead of producing a variety of different images, the generator produces a distribution with single image or just a few variations and the generated samples are almost independent of the latent variable.
We hypothesize that in our case, if the generator occasionally produces an image corresponding to a high-confidence teacher prediction, the cross-entropy loss vanishes and the generator may learn to produce essentially only that output even if other loss components like $\mathcal{L}_\text{M}$ are not fully optimized.
% I think we can provide a better explanation.
Figure~\ref{fig:cifar10-mode-collapse} illustrates a typical example of mode collapse that happened to our generator, where it is able to generate realistic images for the ``automobile'' class but all of the generated objects are red.

As suggested in previous literature \cite{Ghosh2018MADGAN,Liu2016CoGAN}, training multiple generators can be a very simple but powerful way of alleviating this issue.

For our approach we choose to use a setup with $k$ generators, and assign all classes across all generators, so that each class is assigned to exactly one generator. 
Each generator only tries to maximize inceptionism loss for the classes it has assigned. 
For the moment matching, instead of using the moments pre-stored in batch-norm layers, in this case we use a more precise per-category moments for each generator.
The way of estimating these moments is discussed in Section~\ref{sec:exp-CIFAR100}.
% generated moments by generating using DeepDream or a small sample of images\footnote{Note that this data can be collected at the same time teacher is trained, thus preserving the data free nature of distillation}. 

\section{Experiments}
In this section, we turn to an empirical study to evaluate the effectiveness of our method on datasets of increasing size and complexity.
We first conduct a series of ablation studies on CIFAR-10 ($32 \times 32$ image size; $10$ categories) to verify the effect of each optimization objective for generator training,
and then demonstrate the performance of using single or per-class generators on CIFAR-100 ($32 \times 32$ image size; $100$ categories).
Finally, we extend to ImageNet ($224 \times 224$ image size; $1000$ categories) to show how our method scales to larger and more complex datasets.

\subsection{CIFAR-10}

\textbf{Experimental setup.}
The CIFAR-10 dataset \cite{Krizhevsky2009LearningML} consists of $50$K training images and $10$K testing images from $10$ classes.
To make a fair comparison, we follow the setting used in the previous literature \cite{Chen2019DAFL,Fang2019DFAD,Yin2020DeepInversion} with a pre-trained ResNet-34 as the teacher network and ResNet-18 as the student.
The generator architecture is also identical to the one used in~\cite{Chen2019DAFL,Fang2019DFAD}.
The generator is trained using Adam optimizer~\cite{Kingma2014-vr} with a learning rate of $0.001$.
We use the batch size of $256$, $\lambda_\text{S} = 10$, $\lambda_{\ell_2} = 1.5 \times 10^{-5}$, and $\lambda_\text{t} = 6 \times 10^{-3}$.
For CIFAR-10 we only use a single-generator mode. 
More details on the experimental setup can be found in the supplementary materials.

\begin{table}[tbh]
\centering
\small
\begin{tabular}{llccr}
\toprule
Model & Method &  &  & Accuracy \\ \midrule
ResNet-34 & \multicolumn{3}{l}{Supervised Training} & 95.05\%$^\dagger$ \\
ResNet-18 & \multicolumn{3}{l}{Supervised Training} & 93.92\%$^\ddagger$ \\
ResNet-18 & \multicolumn{3}{l}{Knowledge Distillation \cite{Hinton2014KD}} & 94.34\%$^\ddagger$ \\ \midrule
ResNet-18 & \multicolumn{3}{l}{Gaussian Noise} & 11.43\% \\
 & \multicolumn{3}{l}{DAFL \cite{Chen2019DAFL}} & 92.22\% \\
 & \multicolumn{3}{l}{DFAD \cite{Fang2019DFAD}} & 93.3\% \\
 & \multicolumn{3}{l}{Adaptive DeepInversion \cite{Yin2020DeepInversion}} & 93.26\% \\ \midrule
ResNet-18 & Ours & $\mathcal{L}_\text{Inc}$ & $\mathcal{L}_\text{M}$ &  \\
 & \quad $\blackdiamond$ inceptionism & \checkmark &  & 77.31\% \\
 & \quad $\blackdiamond$ moment matching &  & \checkmark & 94.61\% \\
 & \quad $\blackdiamond$ both & \checkmark & \checkmark & \textbf{95.02\%} \\ \bottomrule
\end{tabular}
\caption{
    Test accuracies for different methods on CIFAR-10.
    $^\dagger$The test accuracy of the pre-trained teacher used in this experiment is $95.05\%$, while the ones used in previous literature are slightly stronger: $95.58\%$ \cite{Chen2019DAFL}, $95.5\%$ \cite{Fang2019DFAD} and $95.42\%$ \cite{Yin2020DeepInversion}.
    $^\ddagger$As reported in \cite{Chen2019DAFL}.
}
\label{tab:cifar10}
\end{table}

\begin{figure*}[tb]
    \centering
    \includegraphics[width=0.618\textwidth]{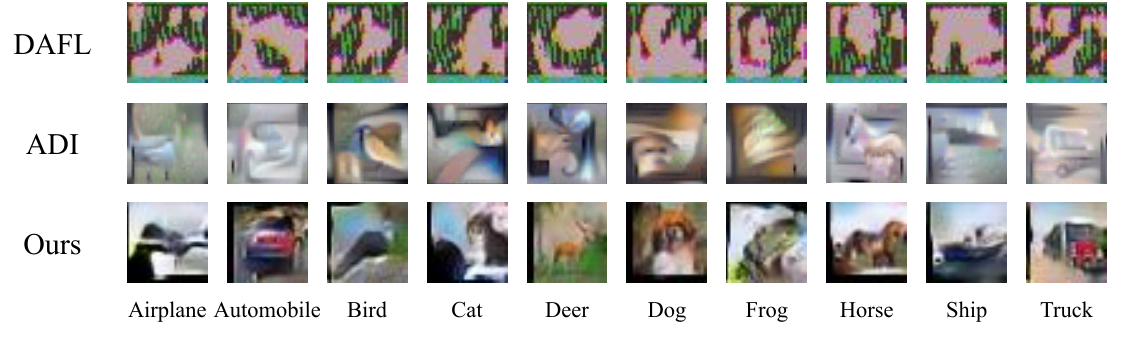}
    \caption{
        From top to bottom, the images generated with DAFL \cite{Chen2019DAFL}, Adaptive DeepInversion (ADI) \cite{Yin2020DeepInversion} and our method, respectively.
        The class labels are listed at the bottom.
    }
    \label{fig:cifar10-visual}
\end{figure*}

\textbf{Experimental results.}
Test accuracies obtained using different methods on CIFAR-10 are summarized in Table~\ref{tab:cifar10}.
Trained in a fully supervised setting, the teacher (ResNet-34) and the student models (ResNet-18) achieve test accuracies of $95.05\%$ and $93.92\%$ respectively.
The student can further gain a $+0.42\%$ improvement by performing knowledge distillation on the teacher using the original training data.
In the data-free setting, as was previously observed, using simple Gaussian noise as input distribution leads to a very poor performance that is only slightly better than a random guess ($10\%$). This is not unexpected since the real data distribution looks very different from Gaussian noise and is generally expected to concentrate on a lower-dimensional manifold embedded in a high-dimensional space.

As part of our ablation study, we first trained the generator using the inceptionism loss $\mathcal{L}_\text{Inc}$ alone.
The resulting student model trained on this generator reached the accuracy of $77.31\%$, which is better than random noise, but significantly lower than the supervised accuracy of $93.92\%$.
In another experiment, where the generator was trained with the moment matching loss $\mathcal{L}_\text{M}$ alone, the resulting student now reached a much higher accuracy of $94.61\%$, 
% \ie only a $?\%$ gap to the supervised performance.
Combining both objectives brings the accuracy of our final method to $95.02\%$, now almost indistinguishable from the accuracy of the original larger teacher model and higher than the accuracy obtained with distillation on the original training dataset.

\textbf{Visualization.}
We finally provide several example images produced by DAFL \cite{Chen2019DAFL}, Adaptive DeepInversion (ADI) \cite{Yin2020DeepInversion} and our generator in Figure~\ref{fig:cifar10-visual}.
As we can see, although ADI can produce images with much higher quality than previous methods, it tends to synthesize images with different textures but similar background (\eg category horse, ship and truck).
In contrast, our method can generate more realistic images, which are likely to have a closer distribution to the original data.

% In Figure \ref{fig:cifar10-visual}, we provides example images generated by DAFL \cite{Chen2019DAFL}, ADI \cite{Yin2020DeepInversion} and our generator for CIFAR-10. We can find that, though ADI works much more better than previous work, it can only generate images with similar background color and slightly different horse ship truck. In contrast, our generator can obtain images that are realistic and much closer to the distribution of the training dataset.

\subsection{CIFAR-100}
\label{sec:exp-CIFAR100}

\textbf{Experiment setup.}
Like CIFAR-10, the CIFAR-100 dataset~\cite{Krizhevsky2009LearningML} also consists of $50$K training images and $10$K testing images, but the images from this dataset are categorized into $100$ classes, which makes it more diverse than CIFAR-10.
In most of our experiments, we use the same model architectures and training hyperparameters as in our CIFAR-10 experiments, with the only exception of $\lambda_\text{S}$ now being chosen as $1$.
More details about the experimental setup can be found in the supplementary materials.

\textbf{Single generator.}
Test accuracies obtained using knowledge distillation with a single generator are listed in Table~\ref{tab:cifar100}.
Our method achieves an accuracy of $76.42\%$ on the test set, which outperforms all previous approaches by a large margin.
However, this result is still slightly worse than the test accuracy of a ResNet-18 network trained in a supervised setting, or distilled from the teacher on the training data.

\begin{table}[t]
\centering
\small
\begin{tabular}{llr}
\toprule
Model & Method & \quad Accuracy \\ \midrule
ResNet-34 & Supervised Training & 77.26\%$^\dagger$ \\
ResNet-18 & Supervised Training & 76.53\%$^\ddagger$ \\
ResNet-18 & Knowledge Distillation \cite{Hinton2014KD} & 76.87\%$^\ddagger$ \\ \midrule
ResNet-18 & Gaussian Noise & 1.23\% \\
 & DAFL \cite{Chen2019DAFL} & 74.47\% \\
 & DFAD \cite{Fang2019DFAD} & 67.7\% \\ \midrule
ResNet-18 & Ours &  \\
 & \quad $\blackdiamond$ single generator & 76.42\% \\
 & \quad $\blackdiamond$ ensembles (meta-data) & \textbf{77.16\%} \\
 & \quad $\blackdiamond$ ensembles (data-free) & 77.02\% \\ \bottomrule
\end{tabular}
\caption{
    Test accuracies of different methods on CIFAR-100.
    $^\dagger$The performance of the pre-trained teacher used in this experiment is $77.26\%$, while the ones used in previous literature are slightly stronger: $77.84\%$ \cite{Chen2019DAFL} and $77.5\%$.
    $^\ddagger$As reported in \cite{Chen2019DAFL}.
}
\label{tab:cifar100}
\end{table}

\textbf{Multiple generators.}
We consider two ways of gathering per-class statistics. 
%In our experiments, we split the whole task of generating all classes of images across multiple generators. For example if we use 10 %generators, we allocate each generator 10 classes. In this setting, instead of computing the moment matching loss with the running %estimates implicitly stored in each batch-norm layer, we use per-class statistics to more precisely instruct a generator to produce images %for a certain category.
One direct way of accumulating it is to: (a) collect a small subset of the training images, (b) feed them to the pre-trained teacher to compute the required moments in each layer, and (c) serve them as meta-data during generator training.
But although we only need a small number of images to gather such statistics, this can no longer be thought as a purely data-free approach.
In the most strict setting, where the training has to be data-free, there is another option in which we can learn several batches of trainable images using Equation~\eqref{eq:img-loss} as the optimization objective \cite{Yin2020DeepInversion,Haroush2020StatMatchQuant}.
Following this approach, we can also obtain a small amount of (synthetic) images to measure  per-class statistics.
Specifically, we sample $100$ images per class from the training data or learn the same amount of images in a data-free manner to compute the per-class statistics for each class, which is then used to train the generators. When performing distillation, we simply sample images uniformly at random from all generators. 

The results of knowledge distillation with the collection of generators are shown in the last two rows in Table~\ref{tab:cifar100}.
Both methods outperform distillation with a single generator and, perhaps more remarkably, a ResNet-18 model trained in a supervised fashion, or the same model distilled on the original dataset.
Finally, we see that both methods exhibit very similar performance, which suggests that we have a freedom to choose a particular approach based on the actual use case.
In a scenario where we can pre-record activation statistics during the teacher training phase, it might be convenient to use the approach relying on meta-data collection.  Otherwise, we may choose the alternative data-free approach without a significant loss of accuracy.

\subsection{ImageNet}

We finally turn to the study on ImageNet \cite{Deng2009ImageNet}. 
\begin{table}[thb]
\centering
\small
\begin{tabular}{lrr}
\toprule
Method & Top-1 Acc. & $\Delta$Acc. \\ \midrule
Supervised Training & 75.45\%(77.26\%$^\dagger$) & N/A \\ \midrule
% Knowledge Distillation \cite{Hinton2014KD} &  &  \\ \midrule
BigGAN \cite{Brock2019BigGAN} & 64.0\%$^\ddagger$ & -13.26\% \\
DeepInversion \cite{Yin2020DeepInversion} & 68.0\%$^\S$ & -9.26\% \\ \midrule
Ours & 69.75\% & -5.70\% \\ \bottomrule
\end{tabular}
\caption{
    Top-1 accuracy of different methods on ImageNet.
    $^\dagger$We use the ResNet-50v1 as our model while DeepInversion \cite{Yin2020DeepInversion} uses ResNet-50v1.5 instead.     
    % , which makes the distillation results not directly comparable.    \notsure{We list the change of accuracy from each one's teacher} at the last column for a clearer comparison.
    $^\ddagger$Reported in~\cite{Yin2020DeepInversion}.
    $^\S$Results obtained without MixUp augmentation.
}
\label{tab:imagenet}
\end{table}

\begin{figure*}[tb]
    \centering
    \includegraphics[width=0.74\textwidth]{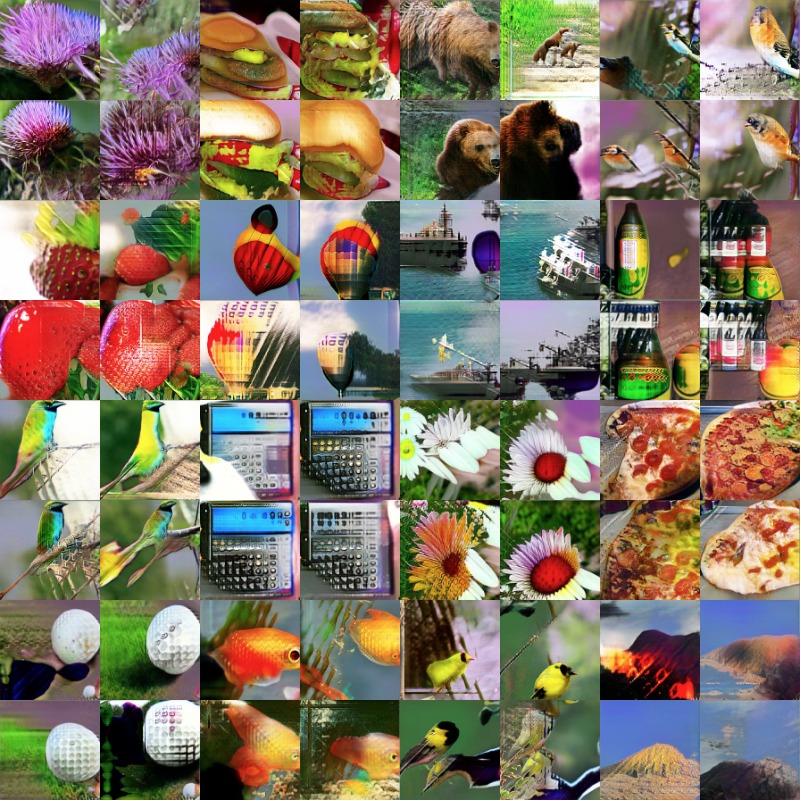}
    \caption{
        Examples of $224 \times 224$ images produced by the generators trained from the given teacher network.
        The teacher is a ResNet-50 classifier trained on ImageNet. 
        We choose $16$ categories suggested by DeepInversion \cite{Yin2020DeepInversion} that images in these categories may have better visual quality.
        We display $4$ images per category in $2 \times 2$ grids.
        Categories from left to right and top to down are: cardoon, cheeseburger, brown bear, brambling, strawberry, balloon, aircraft carrier, beer bottle, bee eater, hand-held computer, daisy, pizza, golf ball, goldfish, goldfinch and volcano.
    }
    \label{fig:imagenet-visual}
\end{figure*}

\textbf{Experimental setup.}
The ImageNet dataset consists of over $1$M images in $1000$ classes.  We explore two target resolutions: $32\times 32$ matching that of CIFAR-10,
and full resolution $224\times 224$. For $32\times32$ images we study the performance trade-offs between the number of generators and the accuracy. For full resolution we train an ensemble of $1000$ generators for distillation, where each generator is trained to produce just a single class. The generator structure is similar to that used in CIFAR-10/100 experiment, we add additional convolutional and $2\times$ upscaling layers to bring the image resolution to the target $224\times 224$. Due to memory constraints we also reduce the number of dimensions of the latent variable $\bm{z}$ from $1024$ to $512$. 
% when putting all the arguments of $1000$ generators once into memory, we have to carefully control the model size.
To estimate per-class statistics we sample $100$ images per class. Note that this sampling can be done during the original teacher training.

All generators are trained with Adam optimizer~\cite{Kingma2014-vr} with a learning rate of $0.001$.
We use the batch size of $64$, $\lambda_\text{S} = 3$, $\lambda_{\ell_2} = 1.5 \times 10^{-5}$, and $\lambda_\text{t} = 6 \times 10^{-3}$.
For the distillation, we use a pre-trained ResNet-50 as the teacher and distill the knowledge to several different students. 
We do not apply any data augmentation techniques such as Inception preprocessing~\cite{Szegedy2016-vu}, MixUp~\cite{zhang2018mixup}, RandAugment~\cite{cubuk2019randaugment} or AutoAugment~\cite{cubuk2019autoaugment} for simplicity and leave further optimization as a subject of future work. More details about training setup is available in the supplementary materials. 

\subsection{Experimental results}
\textbf{Full scale ImageNet.} We illustrate the test results on ImageNet in Table~\ref{tab:imagenet}.
To the best of our knowledge, this is the first  distillation on Imagenet using data-free generative models, and thus no previous work in similar settings that we can directly compare with.  Instead, we display the distillation results using images synthesized by (1) BigGAN~\cite{Brock2019BigGAN}, a generative model trained with a GAN fashion on the ImageNet training data that uses real data to train generators; and (2) DeepInversion~\cite{Yin2020DeepInversion}, which synthesizes images by directly optimizing mini-batches of trainable images one by one. Such kind of method is time-consuming to run but in theory has the potential to produce much more diverse images than using generative models. We note that DeepInversion uses ResNet-50v1.5 as the model and train it with more advanced training techniques, while we are using ResNet-50v1. 
As a result, the performance of our teacher is $1.8\%$ gap to theirs in terms of top-1 accuracy, which makes the distillation results not directly comparable.

Specifically, our student trained with the ensemble of generators achieves an accuracy of $69.75\%$, which outperforms the ones trained with images synthesized by BigGAN and DeepInversion by $5.75\%$ and $1.75\%$ respectively.
Considering their teacher is stronger than us by $1.81\%$, we actually have a even smaller gap ($5.70\%$) to each one's own teacher (BigGAN: $13.26\%$; DeepInversion: $9.26\%$).
% As we clarified, our method is not directly comparable with any methods listed in the table, so these results do not indicate our method is absolutely better than BigGAN or DeepInversion.
% In fact, both of them creates a limited number of images and conduct distillation on that synthetic dataset.
% It is possible that one may choose such methods to use longer time to produce more images and get better distillation results.

%\todo{visualization analysis.}

\begin{table}[htb]
\centering
\small
\begin{tabular}{llrr}
\toprule
Model & \multicolumn{2}{l}{Method} & Top-1 Accuracy \\ \midrule
ResNet-34 & \multicolumn{2}{l}{Supervised Training} & 59.68\% \\
ResNet-18 & \multicolumn{2}{l}{Supervised Training} & 54.99\% \\ \hline
\multicolumn{4}{c}{\cellcolor[HTML]{DDDDDD} Ensemble of different number of generators} \\
ResNet-18 & \multicolumn{2}{l}{Ours} & \\
 & \multicolumn{2}{l}{\quad $\blackdiamond$ \#generators $=1$} & 15.85\% \\
 & \multicolumn{2}{l}{\quad $\blackdiamond$ \#generators $=100$} & 29.40\% \\
 & \multicolumn{2}{l}{\quad $\blackdiamond$ \#generators $=1000$} & 51.82\% \\ \bottomrule
\end{tabular}
\caption{
    Distillation results using ensemble of different number of generators on ImageNet $32 \times 32$.
    The teacher and student are ResNet-34 and ResNet-18 respectively.
}
\label{tab:num-of-generators}
\end{table}

\textbf{Number of generators.}
We further investigate the effect of the number of generators on the accuracy.
Due to the computational and time constraints, we conduct this ablation study on ImageNet resized to $32 \times 32$.
We use ResNet-34 as the teacher and ResNet-18 as the student network.
The single generator is trained with the moving averaged moments stored in each batch-norm layer, and the ensemble of $1000$ generators is trained using per-class statistics.
For the ensemble of $100$ generators, we first divide the image categories into groups of $10$, \ie categories 1--10 in the first group, categories 11--20 in the second group, \etc
Then we sample $1000$ images per group to estimate the per-group statistics for generator training.
The setting of distillation is the same with the experiment on full-resolution ImageNet.

The results are shown in Table~\ref{tab:num-of-generators}.
As we can see, using a single generator we can only obtain a poor distillation accuracy of $15.85\%$. However by using ensembles and increasing the number of generators, we can gradually get better results and finally achieve an accuracy of $51.82\%$, which only has a gap of $3.17\%$ to the supervised-trained student. These results demonstrate the importance of using ensembles to scale generative data-free distillation to large dataset.

\begin{table}[htb]
\centering
\small
\begin{tabular}{lrrr}
\toprule
Student & Sup. Acc. & Distill. Acc. & $\Delta$Acc. \\ \midrule
ResNet-50 & 75.45\% & 69.75\% & -5.70\% \\
ResNet-18 & 68.45\% & 54.66\% & -13.79\% \\
MobileNetV2 \cite{Sandler2018MobileNetV2} & 70.01\% & 43.15\% & -26.86\% \\ \bottomrule
\end{tabular}
\caption{
    Distillation results of students with different architectures on ImageNet.
    The teacher is a ResNet-50 with $75.45\%$ top-1 accuracy.
}
\label{tab:different-students}
\end{table}

\textbf{Different students.}
In our experiments, we also compared the distillation results on students with different architectures (see Table~\ref{tab:different-students}).
Here the teacher is a ResNet-50 model with a top-1 accuracy of $75.45\%$. We use the same set of generators for all students we consider. 
The distillation performance on ResNet-50 is the best with an accuracy drop of $5.70\%$ compared with the model trained in a supervised fashion.
However, the performance results on ResNet\nobreakdash-18 and MobileNetV2 \cite{Sandler2018MobileNetV2} are much worse with larger gaps to their supervised counterparts. This indicates that perhaps there is some entanglement between student and teacher structures that make generators learned on ResNet-50 teacher to be less effective on MobileNetV2 and ResNet\nobreakdash-18 than on ResNet-50 student. 
% without dropping performance too much, and 
Investigation on improving its generalization ability remains subject of future work.
\section{Conclusion}
In this paper, we propose a new method to train a generative image model by leveraging the intrinsic normalization layers' statistics of the trained teacher network.
Our contributions are three-fold.
First, we have shown that the generator trained on our proposed objectives (\ie {\em moment matching loss} and \textit{inceptionism loss}) is able to produce higher-quality and more realistic images than previous methods. Second, we have successfully pushed forward the data-free distillation performance on CIFAR-10 and CIFAR-100 to $95.02\%$ and $77.02\%$ respectively. Finally, we were able to scale it to the ImageNet dataset, which to the best of our knowledge, has not been done using generative models before. 

While we have shown that data-free distillation can successfully scale to large dataset, there are still many open questions. Specifically, our experiments with multi-category generators shows that performance drops dramatically, so generalizing our approaches to work with a single generator is a natural next step. Another direction is to utilize global moments for training images for different classes. Finally, our early experiments show that generators that we produce are teacher specific --- and attempts to use them with a different teachers generally fail. It would be an important extension of our work to create universal generators that allow to learn from any teacher. 
\section*{Acknowledgement}
We thank La\"{e}titia Shao for insightful discussions, Mingxing Tan, Sergey Ioffe, Rui Huang and Shiyin Wang for feedbacks on the draft.

{\small
\bibliographystyle{ieee_fullname}
\bibliography{egbib}
}

\appendix
\clearpage
\section*{Appendix}
\section{Experiment Settings}

We provide more experimental details in this section.
Generally, each generator is trained on an NVIDIA V100 GPU for $10$K steps using the Adam optimizer with $\beta_1=0.9$, $\beta_2=0.999$, $\epsilon=1 \times 10^{-8}$ and a constant learning rate of $0.001$.
We run all the knowledge distillation experiments on a Cloud TPUv3 $32$-core Pod slice.
We use the Momentum optimizer and set the momentum parameter to $0.9$.
We employ a linear warm-up scheme where the learning rate increases from $0$ to the base learning rate for the first $5$K training steps, and then it is decayed by $0.977$ every $1$K steps.
The temperature of distillation is set to $3$.

\subsection{CIFAR-10}

We run knowledge distillation for $60$K steps with a batch size of~$32768$ and base learning rate of~$0.06$.
The generator architecture is illustrated in Table~\ref{tab:generator-arch-cifar}, where $K=10$.

\begin{table}[b]
\small
\centering
\begin{tabular}{c}
\toprule
$z \in \mathbb{R}^{1024} \sim \mathcal{N}(0, I)$ \\
$\operatorname{OneHot}(y) \in \mathbb{R}^K$ \\ \midrule
$\operatorname{Linear}(1024+K) \rightarrow 8 \times 8 \times 128$ \\ \midrule
$\operatorname{Reshape}, \operatorname{BN}, \operatorname{LeakyReLU}$ \\ \midrule
$\operatorname{Upsample} \times 2$ \\ \midrule
$3 \times 3 \operatorname{Conv} 128 \rightarrow 128, \operatorname{BN}, \operatorname{LeakyReLU}$ \\ \midrule
$\operatorname{Upsample} \times 2$ \\ \midrule
$3 \times 3 \operatorname{Conv} 128 \rightarrow 64, \operatorname{BN}, \operatorname{LeakyReLU}$ \\ \midrule
$3 \times 3 \operatorname{Conv} 64 \rightarrow 3, \operatorname{Tanh}$ \\ \bottomrule
\end{tabular}
\caption{
    Generator architecture for CIFAR-10 and CIFAR\nobreakdash-100.
    $K$ denotes the number of classes, where $K=10$ for CIFAR-10 and $K=100$ for CIFAR-100 in single-generator mode; and $K=1$ for the multiple generators experiment.
}
\label{tab:generator-arch-cifar}
\end{table}

\begin{table}[b]
\small
\centering
\begin{tabular}{cl}
\toprule
\multicolumn{2}{c}{$z \in \mathbb{R}^{512} \sim \mathcal{N}(0, I)$} \\ \midrule
\multicolumn{2}{c}{$\operatorname{Linear}(512) \rightarrow 7 \times 7 \times 64$} \\ \midrule
\multicolumn{2}{c}{$\operatorname{Reshape}, \operatorname{BN}, \operatorname{LeakyReLU}$} \\ \midrule
$\operatorname{Upsample} \times 2$ & \multirow{2}{*}{$\times 5$} \\ \cmidrule(r){1-1}
$3 \times 3 \operatorname{Conv} 64 \rightarrow 64, \operatorname{BN}, \operatorname{LeakyReLU}$ &  \\ \midrule
\multicolumn{2}{c}{$3 \times 3 \operatorname{Conv} 64 \rightarrow 3, \operatorname{Tanh}$} \\ \bottomrule
\end{tabular}
\caption{
    Generator architecture for ImageNet.
}
\label{tab:generator-arch-imagenet}
\end{table}

\subsection{CIFAR-100}

We run knowledge distillation for $60$K steps with a batch size of~$32768$ and base learning rate of~$0.1$.
The generator architecture is illustrated in Table~\ref{tab:generator-arch-cifar}.
For the single-generator experiment, $K=100$.
On the other hand, in multiple-generators experiment, each generator is responsible for producing images in a certain category, and therefore $K=1$.

\subsection{ImageNet}

We run knowledge distillation for $300$K steps with a batch size of~$2048$ and base learning rate of~$0.1$.
For the target resolution of $32 \times 32$, we use the same generator architecture as the one used in CIFAR-10/100 experiments.
For the full resolution ($224 \times 224$), each generator is responsible for producing images in one category, whose architecture is illustrated in Table~\ref{tab:generator-arch-imagenet}.

\end{document}